\documentclass[10pt,twocolumn,letterpaper]{article}

\usepackage{wacv}
\usepackage{times}
\usepackage{epsfig}
\usepackage{graphicx}
\usepackage{amsmath}
\usepackage{amssymb}
\usepackage{xcolor}
\usepackage{pifont}
\newcommand{\xmark}{\ding{55}}%
\usepackage{ulem}
\usepackage{booktabs}
\usepackage{tabularx}
\usepackage{float}
\usepackage{amsfonts}
\usepackage{authblk}
\usepackage{capt-of}

\usepackage{footmisc}

\newcommand{\rs}[1]{{\color{red}rs: #1}}

\def\wacvPaperID{0217} 

\wacvalgorithmstrack   

\wacvfinalcopy 

\ifwacvfinal
\usepackage[breaklinks=true,bookmarks=false]{hyperref}
\else
\usepackage[pagebackref=true,breaklinks=true,colorlinks,bookmarks=false]{hyperref}
\fi

\begin{document}

\definecolor{custom_yellow}{HTML}{FFC107}
\definecolor{custom_blue}{HTML}{1E88E5}
\definecolor{custom_jade}{HTML}{00A36C}
\definecolor{custom_lemon}{HTML}{FFC84F}
\newcommand{\blue}[1]{{\color{blue}#1}}

\title{The Change You Want to See}

\author{Ragav Sachdeva}
\author{Andrew Zisserman}
\affil{Visual Geometry Group, Dept.\ of Engineering Science, University of Oxford}

\twocolumn[{
    \vspace{-20pt}
    \renewcommand\twocolumn[1][]{#1}
    \maketitle
    \centering
    \vspace{-10pt}
    \includegraphics[width=0.9\textwidth]{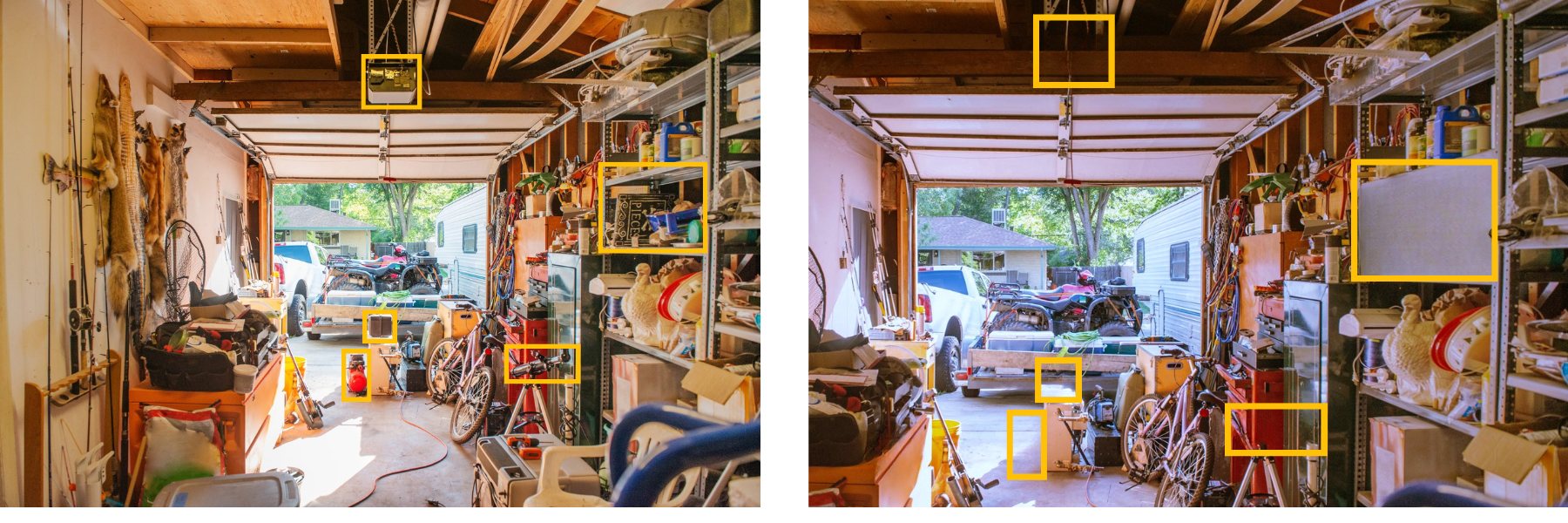}
    \captionof{figure}{In this image pair, 5 out of 6 differences are shown using \textcolor{custom_yellow}{yellow} boxes. Can you spot the remaining one? Our model can.}
    \vspace{20pt}
    \label{fig:teaser}
}]

\maketitle
\thispagestyle{empty}

\begin{abstract}
   \noindent 
We live in a dynamic world where things change all the time. Given two images of the same scene, being able to automatically detect the changes in them has practical applications in a variety of domains. In this paper, we tackle the change detection problem with the goal of detecting ``object-level" changes in an image pair despite differences in their viewpoint and illumination. To this end, we make the following four contributions: (i) we propose a scalable methodology for obtaining a large-scale change detection training dataset by leveraging existing object segmentation benchmarks; (ii) we introduce a co-attention based novel architecture that is able to implicitly determine correspondences between an image pair and find changes in the form of bounding box predictions; (iii) we contribute four evaluation datasets that cover a variety of domains and transformations, including synthetic image changes, real surveillance images of a 3D scene, and synthetic 3D scenes with camera motion; (iv) we evaluate our model on these four datasets and demonstrate zero-shot and beyond training transformation generalization. The code, datasets and pre-trained model can be found at our project page: \blue{\url{https://www.robots.ox.ac.uk/~vgg/research/cyws/}}.
\end{abstract}

\section{Introduction}
Change is all around us. Detecting changes, whether in image pairs or image sequences, is a natural computer vision task. 
Its applications range from the simple ``spot-the-difference" game
to more practical use cases such as in facility monitoring \cite{facility_monitoring},
medical imaging \cite{medical_change}, satellite surveillance \cite{satellite},
and counterfeit detection where, for instance, a forged produce label could have subtle
differences from the original.

The problem we study in this work is the following: given a pair of images,
determine all the changes between them, if any. 
The challenge is to determine the changes between the images that are important
for a particular application, while ignoring ``noise" or ``nuisance" variables
that are irrelevant. For instance, in a surveillance application with a fixed
camera, the ``nuisance" parameters could be the varying lighting of the scene,
changing weather conditions (e.g.\ rain, fog) etc.\ that prevent a simple “difference image” method from being applied.
More generally, the two images may be taken from 
different viewpoints entirely so that in addition to a \textit{photometric}
transformation there maybe also be a \textit{geometric} transformation between them.
Under this setting, determining the differences can also implicitly subsume a
registration problem.

We formulate this problem as the widely studied \textit{detection} problem,
wherein each change is delineated using a bounding box, as opposed to computing
per-pixel changes. This enables ``object-level" change predictions and simplifies
counting the number of changes between two images. To tackle this problem, we introduce a simple Siamese neural network
architecture that operates on two images with geometric and photometric changes and is designed to be
class-agnostic, in that it can detect changes irrespective of the object classes
involved. We make use of an attention mechanism, similar to \cite{coam, Vondrick18}, that can implicitly determine the
correspondences between the images, register them and detect their differences.

To train this architecture, we introduce a scalable method
for generating synthetic training data from real images -- where for
each pair of training images, we know the ground truth bounding boxes
of the differences. The key idea is to leverage existing large-scale image datasets
such as COCO and KITTI, and use off-the-shelf inpainting methods to inpaint various regions in an image to
create differences between the inpainted and the original versions. In addition, we take measures that prevents the model from ``cheating" by
detecting inpainting noise.  Using this dataset, we introduce both
geometric and photometric transformations that we wish to be invariant
to (i.e.\ not important for an application) by standard augmentations
during training.

We demonstrate that a model trained only on this synthetic dataset using affine transformations
and colour jittering can generalize in two significant ways: (i) it can be applied \textit{zero-shot} to other datasets,
and we evaluate its performance over four different datasets including different domains and both real and synthetic
cases; and (ii) the transformations that it can handle extend beyond affine, and we evaluate this by including a dataset with
3D effects due to camera motion. The first generalization is a consequence of using a varied training dataset, 
and the second is a consequence of using attention to determine the correspondences implicitly, rather than
explicitly computing  geometic and photometric transformations between the images.

In summary, we make the following four contributions: (i) we introduce a novel architecture for change detection,
formulated as a detection (rather than segmentation)  problem that is able to implicitly learn correspondences between the images; (ii) we introduce a novel scalable method for generating
a large-scale dataset of training image-pairs from existing object segmentation benchmarks; (iii) we define four
evaluation datasets that cover a variety of domains and transformations: synthetic inpainted COCO image pairs related
by affine transformations; a variety of images with text added in a manner consistent with the geometry of the scene;
real surveillance images of a 3D scene; and synthetic 3D scenes and camera motion using the Kubric pipeline; finally,
(iv) we ablate our design choices, and demonstrate zero-shot and beyond training transformation generalization.


\section{Related Works}
\noindent Since the notion of ``change" is very broad, the problem of exploring changes in a scene has been studied under several different settings. In this section we summarise the contributions of relevant works in each category.\newline

\noindent\textbf{Change captioning:} The change detection problem has been posed as a captioning problem where the model is expected to describe the differences in a pair of images in natural language. Jhamtani et al. \cite{spotthediff} present a Spot-the-difference (STD) dataset of image pairs from surveillance cameras with text based annotations for changes and propose a method that captures visual salience by using a latent variable to align clusters of differing pixels with output sentences. Park et al. \cite{robustchange} focus on semantically relevant changes and present a method that performs robust change captioning on STD as well as a new change detection dataset. Oluwasanmi et al. \cite{captiondiff} propose a fully-convolutional CaptionNet that outperforms previous methods on the STD dataset.\newline

\noindent\textbf{Street-view change segmentation:} Most of the existing works that attempt to localise changes between a pair of images, formulate it as a segmentation problem, particularly in a street-view setting. Sakurada et al. \cite{pcd} proposed a method for segmenting changes in a street scene using a pair of its vehicular, omnidirectional images, with the intention of detecting ``city-scale" changes. Towards the same goal, Alcantarilla et al. \cite{Stent-RSS-16} presented a system for performing structural change detection in street-view videos captured by a vehicle mounted monocular camera over time. Sakurada et al. \cite{weakly} further posed a novel semantic change detection problem and proposed a weakly-supervised silhouette-based model to address it. Recently, Lei et al. \cite{streetscenehei} presented a method to locate the changed regions between a given street-view image pair and demonstrated superior results to the previous methods.

A big challenge faced by these methods is the lack of a large-scale and comprehensively labelled change dataset. Manually labelling all the changed pixels in an uncontrolled setting like streets is an extraordinarily expensive and error-prone task. TSUNAMI and GSV datasets presented in \cite{pcd} contain 100 image pairs each, where the authors report spending 20 minutes to annotate each image pair. PSCD dataset presented in \cite{weakly} contains 500 image pairs where the authors report spending an average of 156 minutes to annotate each image pair. Despite the remarkable effort,  (a) these datasets are relatively small,  and (b) their annotations are not comprehensive (by choice) e.g.\  changes in road signs on ground are not accounted.\newline

\noindent\textbf{Synthetic change detection datasets:} An alternative to collecting and labelling real-world images is to use synthetic datasets where the changes can be controlled. To this end, datasets such as StandardSim\footnote{\label{not_released}These datasets have not been released publicly at the time of writing.} \cite{standardsim} for change detection in retail stores, ChangeSim \cite{changesim} for change detection in warehouses and CARLA-OBJCD\footref{not_released} \cite{carladataset} for change detection in street scenes have been introduced. In this work, we take a tangential approach instead and leverage existing large-scale object detection datasets to train our model. Furthermore, we approach the change detection problem as a bounding box based detection problem (as opposed to segmentation) which makes it possible for us to curate various test sets, with relative ease, to reliably evaluate our model. \newline

\noindent\textbf{Change classification:} Change detection has also been explored as a classification problem by Fujita et al. \cite{change_classification} for damage detection and Wu et al.~\cite{book_change} for detecting changes in book covers.\newline

\noindent\textbf{Correspondence matching:} An orthogonal, yet related, problem to change detection is that of correspondence matching, where the goal is to find corresponding points in images rather than differences. There exists a plethora of literature proposing methods to find corresponding points between a pair of images \cite{Fathy18,Lai20,Savinov17Features,Vondrick18,Wang20,Wang19b, Barnes09,Schoenberger16b}.

\begin{figure*}[ht]
    \centering
    \includegraphics[width=\linewidth]{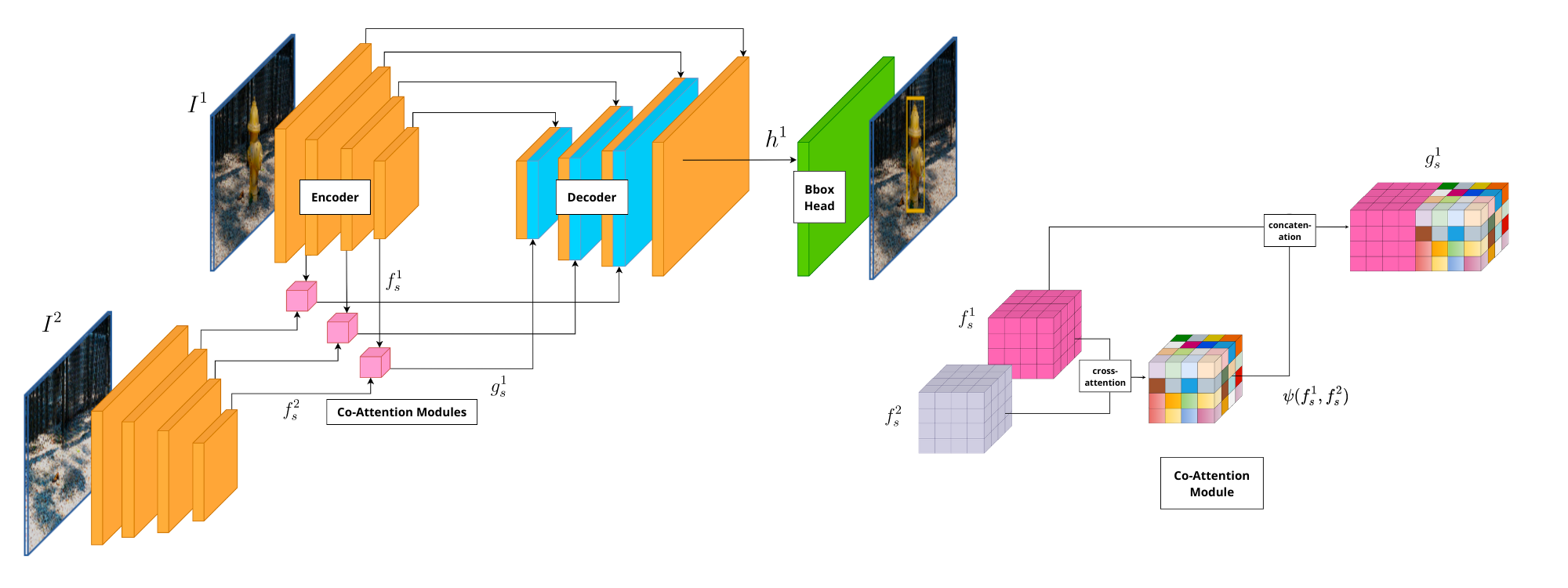}
    \caption{\textbf{Architecture:}  Given two images $I_1, I_2$, an encoder produces feature maps $f_s^1, f_s^2$ respectively at multiple resolutions. A co-attention module is then used to compute conditioned feature maps $g_s^1, g_s^2$ that are impicitly registered with the other image. A U-Net style decoder is then applied to the original and conditioned features maps to produce feature maps $h_1, h_2$. Finally, the bbox detector head uses $h_1, h_2$ to produce bounding boxes for $I_1,I_2$ respectively. For brevity, we only show this pipeline for image $I^1$ (it is symmetric for image $I^2$). Please see Sec.~\ref{sec:network_arch} for details.}
    \label{fig:model_arch}
\end{figure*}

\section{Architecture}
\label{sec:network_arch}
\noindent\textbf{Overview:} 
Given a pair of images under some geometric transformation, our goal
is to localise the changes between them in the form of bounding box
predictions for \textit{each} image. To do so, the model must have a
notion of computing correspondences between the two images and
establishing whether certain regions have changed, while ignoring
nuisance factors such as photometric changes. Therefore, the model
must simultaneously operate on both the images, coalesce their feature
maps in a meaningful way and localise the changed regions.

We achieve this by first obtaining a set of dense feature descriptors
for each image using a CNN-based encoder. These dense feature
descriptors are then conditioned on each other using a co-attention
mechanism that implicitly supplies the correspondences. Next, these conditioned feature descriptors are passed
through a decoder to obtain high resolution conditioned image
descriptors which are used by a bounding box detection head to
localise the changes.  Briefly, we employ a siamese architecture
comprising of a U-Net model \cite{unet}, modulated with co-attention
layers~\cite{coam} and concurrent Spatial and Channel Squeeze \&
Excitation blocks (scSE) \cite{scse}, followed by a bounding box
prediction head \cite{centernet}, as shown in
Fig.~\ref{fig:model_arch}.\newline

\noindent In detail, given two images $I^1 \in \mathbb{R}^{3\times H\times W}$ and $I^2 \in \mathbb{R}^{3\times H\times W}$ with an unknown geometric transformation between them, the model used to localise changes between them is split into four components.\newline

\noindent\textbf{U-Net Encoder:} First, we encode $I^1, I^2$ using a U-Net encoder (CNN), represented by $\Phi_{E}(\cdot)$, to obtain dense feature descriptors at multiple spatial resolutions $s$. Specifically, we obtain feature maps $f^1_s \in \mathbb{R}^{c_s \times h_s \times w_s}$ and $f^2_s \in \mathbb{R}^{c_s \times h_s \times w_s}$ for images $I^1, I^2$ respectively, where $s \in \{1,2,3\}$, after the last three blocks in a ResNet50 \cite{resnet} model.\newline

\noindent\textbf{Co-Attention Module:} In order to predict changed regions in $I_1$, its feature maps must also embed information from $I_2$, and vice versa. We, therefore, wish to propagate the information embedded in $f^1_s$ and $f^2_s$ to each other in order to facilitate the computation of what has ``changed". To permit this information exchange, we make use of the co-attention module~\cite{coam}. Intuitively, each feature vector at location $(x^1,y^1)$ in $f^1_s$ attends to feature vectors at all locations $(x^2,y^2)$ in $f^2_s$ and is concatenated to their weighted sum (and vice-versa). This can be thought of as spatially warping the feature vectors of one image and concatenating with the other such that the two images are \textit{registered}. Formally, we obtain the co-attended features $g^1_s = [f^1_s \mathbin\Vert \psi(f^1_s, f^2_s)]$ and $g^2_s = [f^2_s \mathbin\Vert \psi(f^2_s, f^1_s)]$, where $[\ \mathbin\Vert\ ]$ is the concatenation operation (along channel $c$) and $\psi(\cdot)$ is the cross-attention mechanism defined as

\begin{equation}
\label{eq:coam}
\psi(f^q,f^k)_{cij} = \sum\limits_{l}\sum\limits_{m}A_{ijlm}.V_{clm}
\end{equation}
where,
\begin{equation}
\label{eq:attention}
A_{ijlm} = \text{Softmax}(\sum\limits_{c}Q_{cij}.K_{clm}, \text{dim=}l,m)
\end{equation}
and,
\begin{equation}
Q = \mathbf{W}^qf^q,\ K = \mathbf{W}^kf^k,\ V = f^k
\end{equation}
where $\mathbf{W}^q$ and $\mathbf{W}^k$ are learnable parameters. Thus, the feature maps $g^1_s$ and $g^2_s$ are conditioned on \textit{both} the images and contain adequate information to localise the changes.\newline

\noindent\textbf{U-Net Decoder:} Following this, we upsample and decode $g^1_s$, $g^2_s$ using the U-Net decoder (with skip connections from the encoder), modulated with scSE blocks \cite{scse}, represented by $\Phi_D(\cdot)$ to produce feature maps $h^1$ and $h^2$ respectively, at the original image resolution.\newline

\noindent\textbf{Bbox Head:} Finally, $h^1$ and $h^2$ are fed into a CenterNet head, which minimises the detection loss function as described in \cite{centernet}, to produce bounding boxes around changed regions in both the images.

\begin{figure*}[ht]
    \centering
    \includegraphics[width=\linewidth]{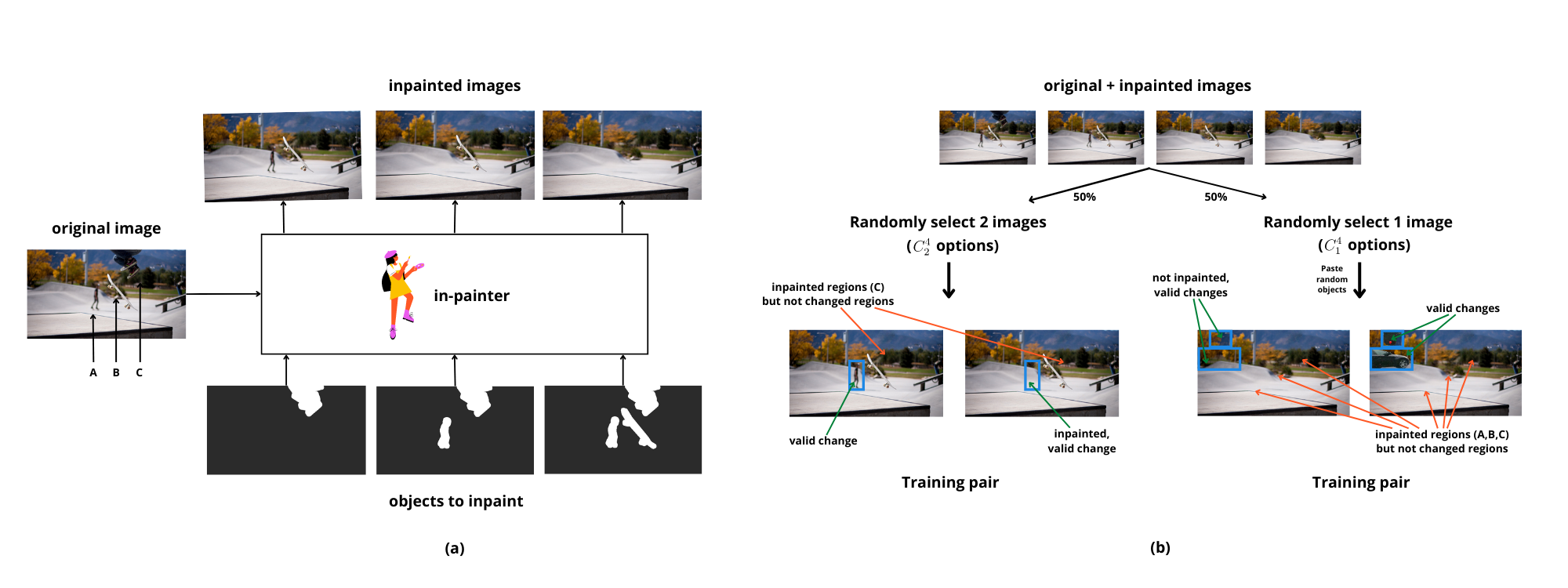}
    \caption{\textbf{Training data generation pipeline:} The figure above illustrates how we generate several image pairs with changed regions from a single COCO image. Given an original COCO image, we first (a) use an in-painting method to compute several images with inpainted regions. Then (b) given the original image along with inpainted images, we randomly sample an image pair for training, along with their ground truth \textcolor{custom_blue}{bounding boxes}, as shown. Notice that the image pairs can have inpainted regions that are not valid changes. This prevents the model from collapsing to simply learning inpainitng noise patterns. Please see Sec.~\ref{sec:inpainting} for details.}
    \label{fig:inpainting_pipeline}
\end{figure*}

\section{No Change Detection Dataset? No Problem}
\label{sec:inpainting}
\noindent Much of the recent success of deep learning methods is attributed to the availability of large-scale training datasets with reliable annotations. Currently, however, there are no publicly available datasets for the change detection problem, as formulated in this work. To avoid curating and manually labelling thousands of image pairs with changes, we propose a procedure to leverage existing large-scale image datasets and state-of-the-art image inpainting methods to simulate visually realistic ``changes". The complete training data processing pipeline is shown in Fig.~\ref{fig:inpainting_pipeline} while we delineate the details below.\newline

\noindent\textbf{Inpainted changes:} For this work, we make use of the COCO dataset \cite{Lin2014MicrosoftCC}, which comes with bounding boxes and segmentation masks for various objects in each image. Given a COCO image along with a binary segmentation mask of various objects in it, we inpaint the said objects using a state-of-the-art image inpainting method, LaMa \cite{lama}, to make the objects ``disappear". The resulting inpainted image along with the original COCO image now constitute an image pair with changes (disappeared objects), for which we have the ground truth annotations (bounding boxes for the objects in the original COCO image).\newline

\noindent\textbf{Combating inpainting noise:} Even though inpainting produces seemingly realistic changes, we noticed that the inpainted regions tend to have ``noise" (as observed by other works in the literature \cite{noise_doesnt_lie, forgery_detection}). In order to discourage the model from simply learning this inpainting noise instead of learning the actual changes between images, we adopt the following two strategies:
\begin{itemize}
    \item For each COCO image we obtain multiple inpainted images, each with a different subset of inpainted objects, out of which we randomly sample two. For instance, consider an image with 3 objects: A, B \& C. Let's say we obtain two inpainted images: $I_1$ which only has object A (B \& C have been inpainted), and $I_2$ which only has object B (A \& C have been inpainted). In this case the model must predict two bounding boxes per image (B does not exist in $I_1$ and A does not exist in $I_2$, hence two changed regions per image). At the same time, the model must learn to ignore inpainting noise for C which has disappeared in both the images and is therefore not a valid change. Thus we force the model to learn actual visually-salient changes.

    \item Aside from inpainted changes, we also ``paste" random objects into the image (taken from a different random COCO image) to simulate changes. While these inserted objects seem visually unrealistic, it requires the model to predict the ``missing" object in the original image, which does not have any inpainting noise.
\end{itemize}

\noindent\textbf{Training dataset:} We randomly select 60000 images from the COCO train subset as our ``original" images. For each original image, we use LaMa \cite{lama} to generate $n \in \{1,2,3\}$ images, each with a different subset of objects inpainted, as described in the pipeline above. We then randomly split these 60000 samples (each with $C^{n+1}_{2}$ image pairs) into training and validation set consisting of 57000 and 3000 samples respectively. Each image is resized to $256\times256$ pixels (due to computational constraints), along with the appropriate scaling of its ground truth change bounding boxes. Given an image pair, we apply random affine transformations (scale $\in [0.8, 1.5]$, translation $\in [-0.2,0.2]$ and rotation $\in [-\frac{\pi}{6}, \frac{\pi}{6}]$) to each image independently and adjust the ground truth bounding boxes appropriately. In addition, we apply random colour jittering to make our model invariant to photometric changes.
We note that the change annotations are class agnostic in that they do not have access to the COCO class labels, rather the only classification is that something has changed at the scale of the bounding box. The validation set is strictly used to pick the best model (with the lowest loss) for evaluation and does not inform the training in any other way.

\begin{figure*}[ht]
    \centering
    \includegraphics[width=\linewidth]{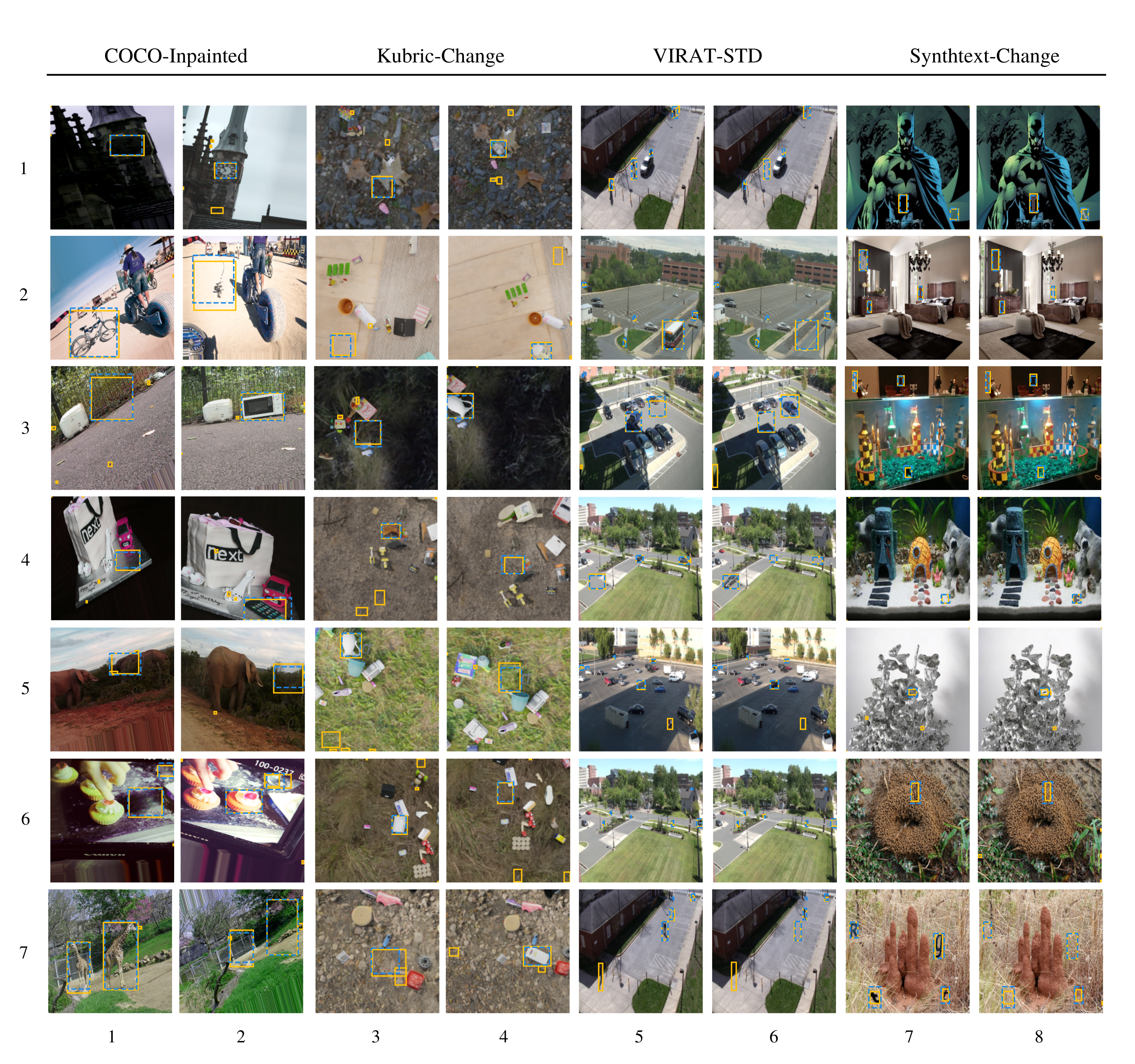}
    \caption{\textbf{Qualitative results:} We show the bounding box \textcolor{custom_yellow}{predictions (solid)} of our model on all the test sets, along with the \textcolor{custom_blue}{ground truth (dashed)}. Since the detection head outputs 100 bounding boxes per image (see Sec.~\ref{sec:implementation}), for the purpose of visualisation, we display the 5 most confident predictions. In case of multiple bounding boxes with significant overlap, we keep the most confident and suppress the others. 
    Note the significant photometric changes in COCO-Inpainted, 3D geometric effects in Kubric-Change (notice the inside of the cup in row 2, col 3-4), detection of really small objects in VIRAT-STD (even picking up valid changes that are not part of the ground truth e.g. row 5, col 5-6) and very subtle letters in Synthtext-Change. We recommend that the reader zooms in on the individual image pairs for inspection.}
    \label{fig:predictions}
\end{figure*}

\begin{figure*}[ht]
    \centering
    \includegraphics[width=\linewidth]{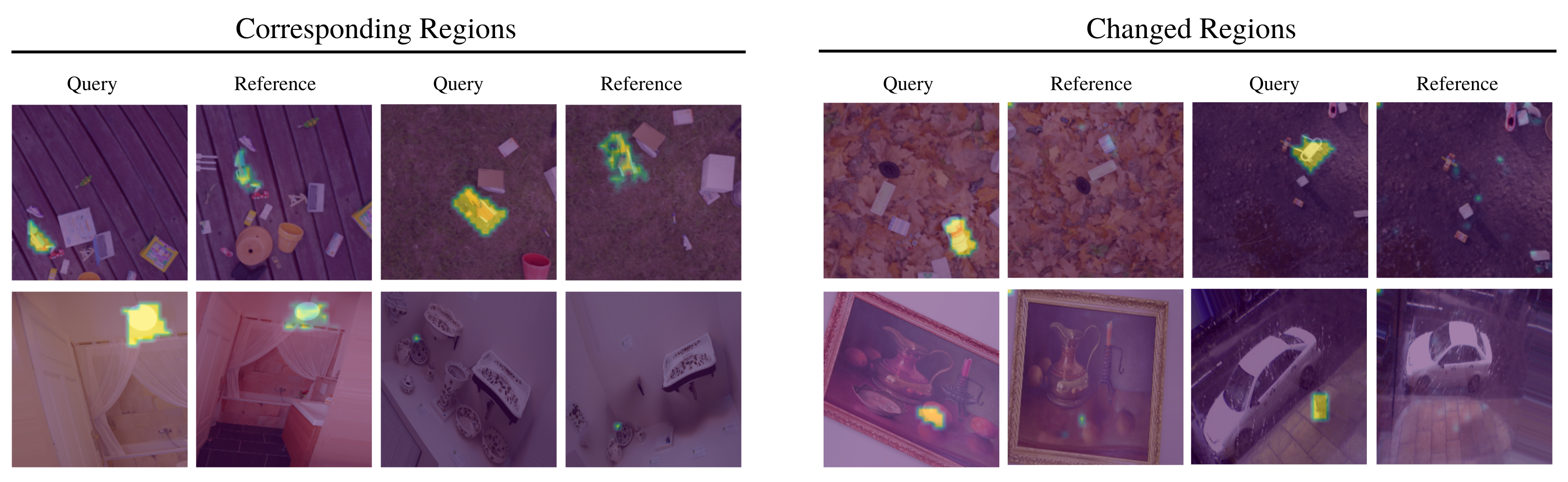}
    \caption{\textbf{Co-attention maps:} Given some pre-determined regions in $I_1$ (\textsc{Query}), we visualise the cross-attended regions in $I_2$ (\textsc{Reference}) from the (spatially highest resolution) \textsc{coam} layer of our model. The examples on the left show varying shape and size of \textsc{Query} regions, including the correspondence for a single pixel. The examples on the right show cases where the query region selected has no correspondence on the right (as the object is missing) -- in these cases the attention map correctly does not highlight a region in the \textsc{Reference} images. It is evident that the model has not only learnt to establish corresponding regions between the two images, but also fine-grained point-to-point correspondences.}
    \label{fig:attention}
\end{figure*}

\begin{table*}[ht]
\footnotesize
\centering
\begin{tabular*}{\linewidth}{c @{\extracolsep{\fill}} ccccccccc}
\toprule
backbone & \# attn modules & attn type & scSE &  geom transformation &  \multicolumn{4}{c}{coco-inpainted test set (AP)} \\
&&&& (train \& test) & small & medium & large & all \\
\midrule
ResNet18 & 2 & \textcolor{custom_jade}{\textsc{coam}} & \xmark & affine & 0.08 & 0.16 & 0.26 & 0.11 \\
ResNet18 & 3 & \textcolor{custom_jade}{\textsc{coam}} & \xmark & affine & 0.32 & 0.49 & 0.49 & 0.37 \\
ResNet50 & 3 & \textsc{noam} & \xmark & affine & 0.15 & 0.32 & 0.49 & 0.21 \\
ResNet50 & 3 & \textcolor{custom_jade}{\textsc{coam}} & \xmark & affine & \textbf{0.46} & 0.74 & 0.70 & 0.58 \\
ResNet50 & 3 & \textcolor{custom_jade}{\textsc{coam}} & \checkmark & affine & \textbf{0.46} & \textbf{0.79} & \textbf{0.85} & \textbf{0.63} \\
\midrule
\midrule
ResNet50 & 3 & \textcolor{custom_jade}{\textsc{coam}} & \checkmark & identity & 0.60 & 0.89 & 0.94 & 0.73 \\
ResNet50 & 3 & \textsc{noam} & \checkmark & identity & \textbf{0.68} & \textbf{0.93} & \textbf{0.95} & \textbf{0.79} \\
\bottomrule
\\
\end{tabular*}
\caption{\textbf{Ablation study:} We ablate various components of our model and report the AP on two variants (\textit{affine}, \textit{identity}) of the COCO-inpainted test set. Note that due to out-of-bounds cropping when applying geometric transformations, \textit{affine} and \textit{identity} test sets do not necessarily have the same number of changes,  and methods trained and tested on one should not be directly compared with the other.}
\label{tab:ablation}
\end{table*}

\begin{table*}[ht]
\footnotesize
\centering
\begin{tabular*}{\linewidth}{c @{\extracolsep{\fill}} |cccc}
\toprule
test set & COCO-Inpainted & Synthtext-Change & VIRAT-STD & Kubric-Change  \\
\midrule
type & inpainting & synthetic & real & photo-realistic sim \\
fixed camera & \checkmark & \checkmark & \checkmark & \xmark \\
geometric transformation \rule{0pt}{1ex} & Affine & None & None & 3D \\
\# image pairs & 4408 & 5000 & 1000 & 1605 \\
\midrule
result (AP) & 0.63 & 0.89 & 0.54 & 0.76 \\
\bottomrule
\newline
\end{tabular*}
\caption{\textbf{Quantitative results:} We report the AP of our model (ResNet50, 3 \textsc{coam} layers, with scSE blocks) on various test sets.}
\label{tab:results}
\end{table*}

\section{Experiments}
\noindent Given two images, under some geometric transformation from one another, we aim to localise the changed regions while being invariant to photometric changes. This section describes the datasets we used to test our model and various implementation details, along with the results.

\subsection{Evaluation datasets}
\label{sec:testing_datasets}
\noindent To evaluate the performance of our model, we contribute four test datasets as described below. Please see Fig.~\ref{fig:predictions} for example image pairs.\newline

\noindent\textbf{COCO-Inpainted:} We curate an inpainting-based test set from the COCO test subset. We divide this test set into 3 categories based on the size (\textit{small}, \textit{medium} and \textit{large}, as defined in \cite{coco_evaluation}) of the inpainted objects. Using the same methodology as described in Sec.~\ref{sec:inpainting}, we curate $1655$ image pairs for \textit{small}, $1747$ image pairs for \textit{medium} and $1006$ image pairs for \textit{large}, giving us a total of $4408$ image pairs for this test set. Furthermore, we apply random affine transformations to the images along with colour jittering. Due to the affine transformations and cropping there will be some regions of the image that have no correspondence in the other image. Please see the first example pair in Fig.~\ref{fig:predictions} for reference. \newline

\noindent\textbf{Synthtext-Change:} We use the pipeline described in \cite{synthtext} to synthetically add random text to ``background" images and generate $5000$ image pairs with text-based changes in a manner that is consistent with their geometry. We do not augment the images any further i.e. the images have an identity geometric and photometric transformation. Note that in order to simplify quantitative evaluation, the generated texts are reasonably-spaced letters of varying sizes. This avoids having to deal with letter-level, word-level and paragraph-level predictions, where the model groups spatially-close small letters into a single bounding box but predicts a bounding box for each letter for bigger font-sizes.\newline

\noindent\textbf{VIRAT-STD:} To detect outdoor scene changes, we select $1000$ image pairs at random from the STD dataset \cite{spotthediff}. These image pairs are originally taken from the VIRAT Video Dataset \cite{virat}, which has bounding box annotations for several objects in each video frame. Since STD does not come with ground truth bounding box annotations for changes, we use a best-effort automated pipeline to obtain the ground truth (with a small percentage of them manually verified by human-in-the-loop). Since the camera is static, there is an identity geometric
transformation between the images (though there may be small motions of the camera due to wind etc.), but the photometric conditions may change due to time-of-day, weather conditions etc. \newline

\noindent\textbf{Kubric-Change:} We use the recently introduced Kubric dataset generator \cite{kubric} to curate $1605$ realistic-looking image pairs with controlled changes. The scenes consist of a randomly selected set of 3D objects resting on a randomly textured ground plane. For a given scene, we iteratively remove objects from it and capture ``before" and ``after" image pairs.
Unlike the datasets above where there is a planar geometric
transformation between the images (affine or identity), for these
image pairs the camera centre moves. Since the scene is 3D there can be parallax and occlusion/disocclusion
changes between the image pairs.\newline

\subsection{Implementation}
\label{sec:implementation}
\noindent We use ResNet50 \cite{resnet} as the encoder for our U-Net model (with ImageNet pre-trained weights), with 5 blocks (1-5), where we apply the co-attention module to the feature maps of blocks 3-5. The U-Net decoder also has 5 blocks with depths $(256, 256, 128, 128, 64)$, along with scSE blocks \cite{scse}. The CenterNet head is implemented as described in \cite{centernet} with the hidden channel dimension of $64$ and is configured to predict $100$ bounding boxes per image. The overall model has 49.5M trainable parameters and is trained on $2$ P$40$ GPUs for $200$ epochs, using the DDP training strategy with a batch size of $16$. We use Adam \cite{adam} to optimise the overall objective with learning rate of $0.0001$ and weight decay of $0.0005$.

\subsection{Evaluation Metrics}
\noindent To quantitatively evaluate our model, we compute the Average Precision (AP) metric defined in \cite{pascalvoc}, as is standard. We emphasise the fact that for each image pair, the model outputs bounding boxes of changed regions for \textit{both} the images and is evaluated as such.

\subsection{Ablation}
\noindent To study the effect of various modules of our method, we ablate different components of our model and show its performance on the 3 subsets (\textit{small}, \textit{medium} and \textit{large}) of the COCO-Inpainted test set. As evident from Table~\ref{tab:ablation}, using more attention modules (3 instead of 2), using a bigger model (ResNet50 as opposed to ResNet18) and adding scSE blocks \cite{scse} all lead to an improvement in the results. 

Furthermore, given two images under affine transformation, we recognise that it is possible to register them if their transformation matrix is known. Consequently, if we know a priori that the images are registered, we note that it is possible to replace the co-attention module (\textsc{coam}) with a simpler module, which we call no-attention module (\textsc{noam}), which simply concatenates the features maps from the two images i.e. $\psi(f^q,f^k) = f^k$ in eq.~\ref{eq:coam}. The results from Table~\ref{tab:ablation} show that \textsc{coam} is almost on par with \textsc{noam} when the images are under an identity transformation, however, \textsc{noam} is much worse than \textsc{coam} under geometric transformations.

\subsection{Results}
\noindent We take our model (ResNet50 backbone, 3 \textsc{coam} layers, with scSE blocks) trained on the dataset described in Sec.~\ref{sec:inpainting} (with affine transformation) and evaluate it on $4$ test sets without any further training/finetuning. We show some qualitative predictions by the model in Fig.~\ref{fig:predictions} on each of the test sets and report the average precision values in Table~\ref{tab:results}. To the best of our knowledge, there are no existing works that tackle the change detection problem using a bounding-box based method which makes it difficult to compare our method with prior-art.

Our results show that not only is our method able to detect changes under extreme affine transformation and colour-jittering for the COCO-Inpainted test set, but also it is able to generalise zero-shot to changed image pairs procured from very different data distributions. Particularly, we note that even though our model is only trained using affine transformations, it produces impressive results on the Kubric-Change test set, where the changed image pairs are no longer related by a homography due to the movement of camera centre and the fact that the objects in the scene are 3D.

In Fig.~\ref{fig:attention} we show the visualisation of attention maps from the co-attention module. Specifically, given feature maps $f_s^1 \in \mathbb{R}^{C\times I\times J}$ and $f_s^2 \in \mathbb{R}^{C\times L\times M}$ of images $I_1$ and $I_2$ respectively, we obtain $A \in \mathbb{R}^{I\times J\times L\times M}$ using eq. \ref{eq:attention}. Then, for a set of query locations
$q$ in $f_s^1$, the co-attention map $G$, given by
\begin{equation}
G_{lm} = \max\limits_{(i,j) \in q} A_{ijlm},
\end{equation}
represents the attended locations in $f_s^2$.
It is evident from the visualisations that the model has learnt to establish correspondences between the two images, which is a logical step towards finding the changes.

\section{Conclusion}
\noindent Humans have a hard time finding changes in a scene -- which is why we tend to find ``spot the difference" tasks to be quite challenging. Adding viewpoint and photometric changes on top of this already difficult problem further elevates its perplexity. In this work we tackled the problem of automatically detecting changes in two images of the same scene under some geometric transformation, while ignoring nuisance factors such as photometric changes. We study a new formulation of this problem and treat it as a bounding-box based detection problem. Due to the lack of a large-scale training dataset for this problem, we proposed a training data generation pipeline that leverages existing datasets (or any arbitrary collection of images for that matter) and off-the-shelf image inpainting methods. Finally, we proposed and trained a novel neural network (in an end-to-end manner using a standard detection loss \cite{centernet}) and showed that it is able to successfully zero-shot detect changes (without any finetuning or sim2real training) on several new benchmarks.
\\[\baselineskip]
\noindent \textbf{Limitations:} The method proposed in this work largely focuses on the detection of changes in \textit{things} rather than \textit{stuff} (as defined in \cite{caesar2018coco}). While it is likely that the trained model has the capacity to detect \textit{stuff} changes, we have not investigated this.
Furthermore, since the trained model is a \textit{change} detector and not an \textit{object} detector, it may group several overlapping changed objects into a single bounding box (as a single changed ``object'').
Finally, due to the nature of the training data, the model was largely tested on relatively planar scenes with mild occlusions/dis-occlusions. Going beyond, to general two-view scenes, with significant changes in the camera pose and challenges such as parallax and severe occlusions/dis-occlusions is a natural direction for future work.
\\[\baselineskip]
\noindent \textbf{Acknowledgements:} We would like to thank Charig Yang, Laurynas Karazija, Luke Melas-Kyriazi, Aleksandarv (Suny) Shtedritski and Yash Bhalgat for proof-reading the paper. This research is supported by EPSRC Programme Grant VisualAI EP/T028572/1 and a Royal Society Research Professorship RP\textbackslash R1\textbackslash191132.

{\small
\bibliographystyle{ieee_fullname}
\bibliography{egbib,shortstrings,olivia}
}

\end{document}


\title{Supplementary Material}

\maketitle
\thispagestyle{empty}

\section{Overview}
In this supplementary material we include a few more details that we omitted in the original paper.
\begin{itemize}
    \item In Sec.~\ref{sec:kubric_more} we talk about the pipeline used to curate the Kubric Change dataset along with some visualisations showing the change in camera positions for ``before" and ``after" images.
    \item In Sec.~\ref{sec:aff_to_proj} we show our method's predictions on projective transformations.
    \item In Sec.~\ref{sec:examples} we show more qualitative examples for each of the test datasets.
\end{itemize}

\section{More on Kubric-Change}
\label{sec:kubric_more}
\begin{figure}[H]%
    \centering
    \subfloat[\centering]{{\includegraphics[width=0.4\columnwidth]{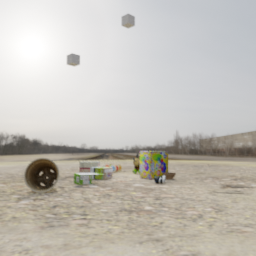} }}%
    \subfloat[\centering]{{\includegraphics[width=0.4\columnwidth]{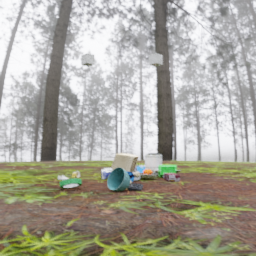} }}%
    \caption{The cubes ``floating" in air represent the before and after camera positions.}%
    \label{fig:kubric_cameras}%
\end{figure}

In this section we briefly describe the pipeline used to acquire the Kubric-Change dataset. We build upon the \href{https://github.com/google-research/kubric/blob/main/challenges/movi/movi_def_worker.py}{\blue{movi\_d}} script provided by Kubric authors, which selects $n \in [10,20]$ random objects out of $1000+$ assets and spawns them into a random scene at random locations bounded by [(-7, -7, 0), (7, 7, 10)]. It then runs a physics simulator for 100 frames for the objects to fall and settle. We then spawn the camera randomly in a cuboid bounded by [(-5,-5,12), (5,5,18)] and take a picture of the scene. We then remove the ``most visible" object and re-spawn the camera randomly in the cuboid at maximum distance of $7$ and maximum rotation of $\frac{\pi}{6}$ from its previous pose. We then take another picture. This process is repeated multiple times to collect a large enough dataset. Fig.~\ref{fig:kubric_cameras} shows two examples of before and after camera configurations to help put things into perspective.

\begin{figure}[H]%
    \centering
    \subfloat[\centering]{{\includegraphics[width=0.4\linewidth]{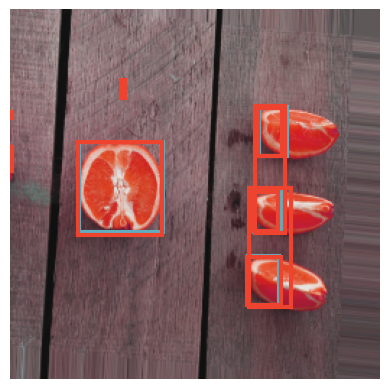} }}%
    \subfloat[\centering]{{\includegraphics[width=0.4\columnwidth]{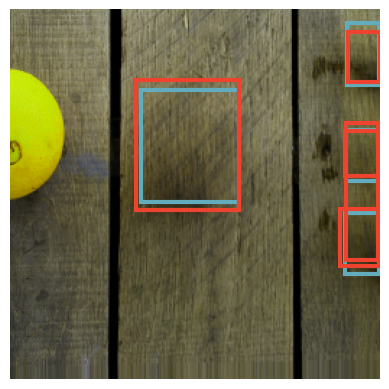} }}%
    \caption{Ground truth annotations in blue. The right half of the three (vertically arranged) orange slices in (a) is not visible in (b) so we cannot be sure whether it has changed or not. Consequently, only the left half is annotated in the ground truth.}%
    \label{fig:partial_bbox}%
\end{figure}

In addition, we make sure that the changed annotations are only over the regions visible in both the images. 
Fig.~\ref{fig:partial_bbox} illustrates this. Even though the orange slices have disappeared, the ground truth annotations (blue boxes) are only over the regions which are visible in both the images.

\section{Affine to Projective Generalisation}
\label{sec:aff_to_proj}
While it was not the focus of this work, we show that our model is also able to generalise to projective transformations while having only been trained using affine transformations. We have no doubt that an explicit training on projective transformations will result in even better predictions.

\begin{figure}[H]%
    \centering
    \subfloat[\centering]{{\includegraphics[width=\columnwidth]{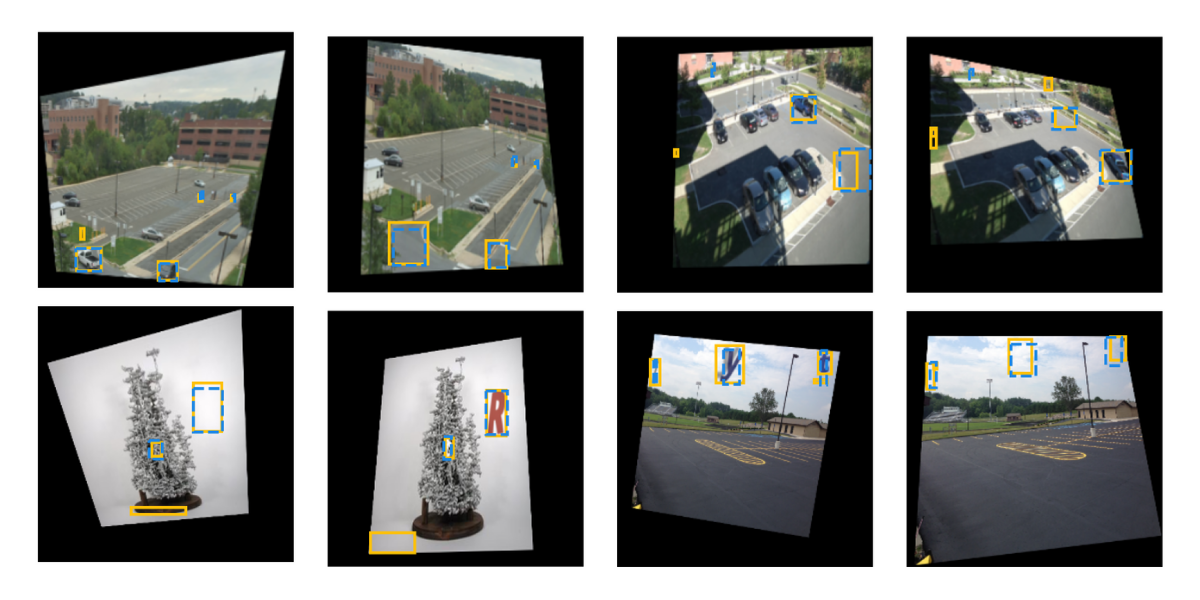} }}%
    \caption{Predictions of our model on images related by a projective transformation.}%
    \label{fig:projective}%
\end{figure}

\section{More dataset examples}
\label{sec:examples}
In this section we show more examples for the four test sets presented in this paper. The first two columns show Image 1 and Image 2 that are fed into the model. The last two columns show the top-5 predicted bounding boxes (in yellow, solid), suppressing the ones with significant overlap, and the ground truth (in blue, dashed). Please find the figures below.

\begin{figure*}[ht]
    \centering
    \includegraphics[width=0.9\linewidth]{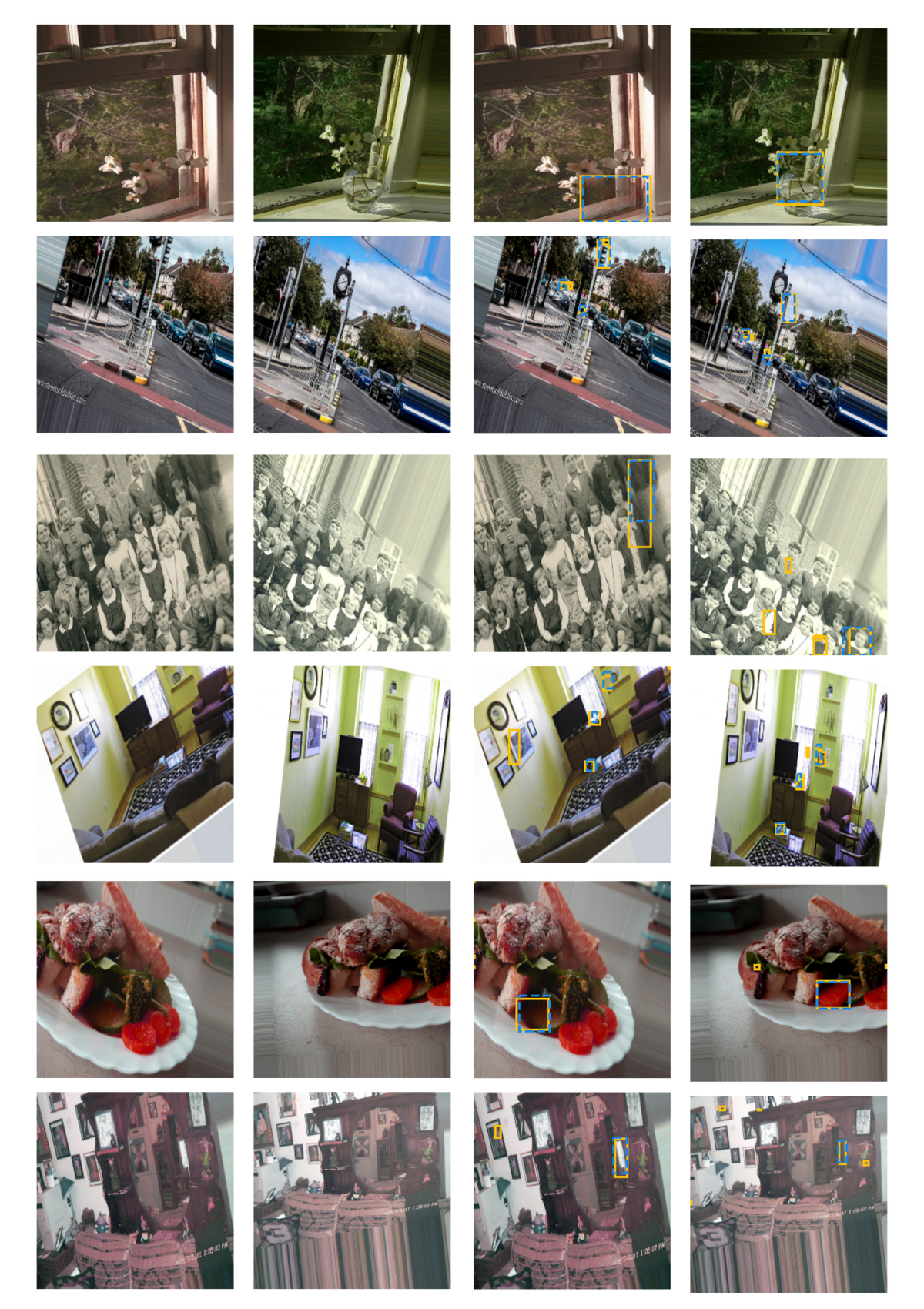}
    \caption{COCO-Inpainted}
    \label{fig:coco}
\end{figure*}

\begin{figure*}[ht]
    \centering
    \includegraphics[width=0.9\linewidth]{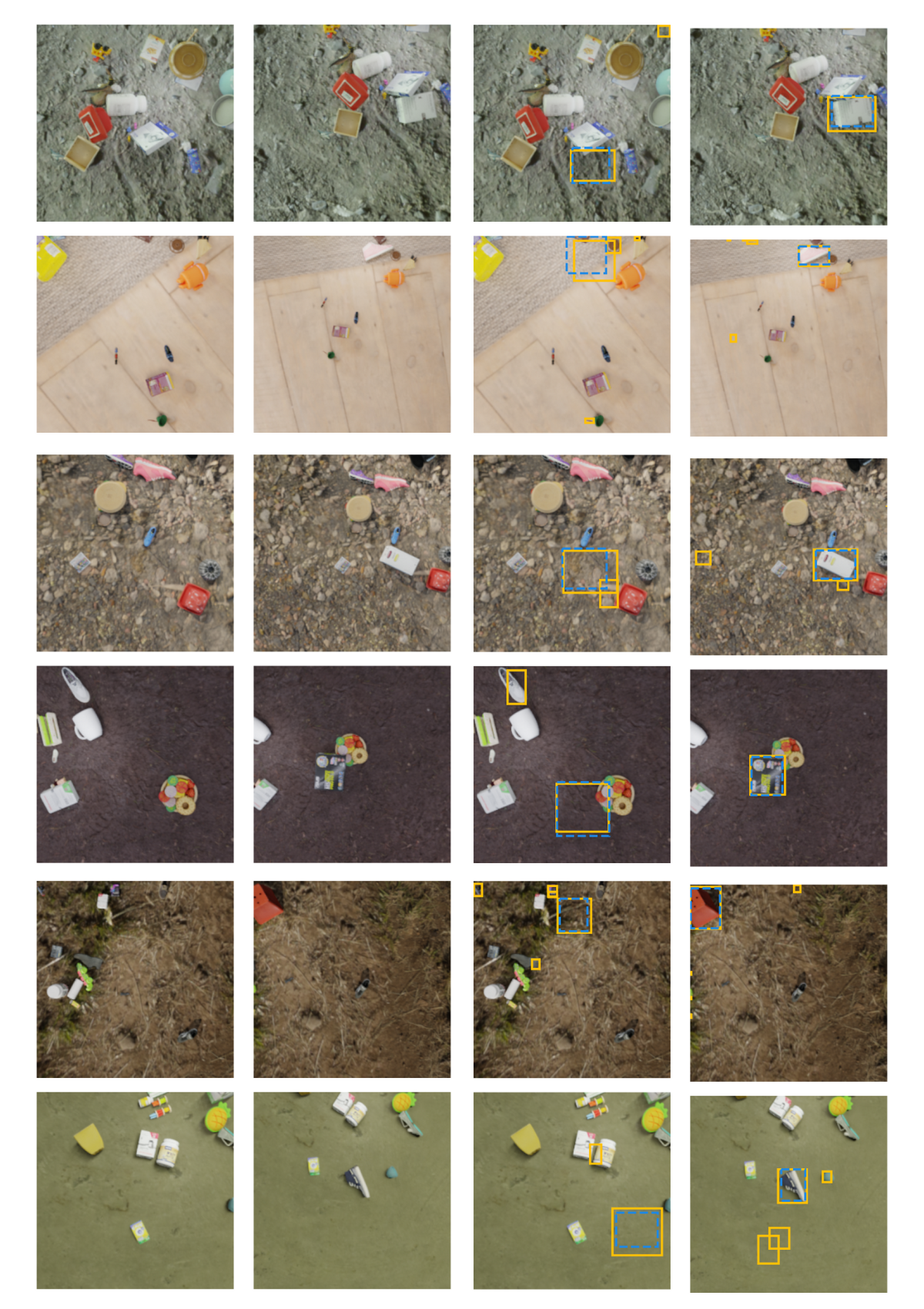}
    \caption{Kubric-Change}
    \label{fig:coco}
\end{figure*}

\begin{figure*}[ht]
    \centering
    \includegraphics[width=0.9\linewidth]{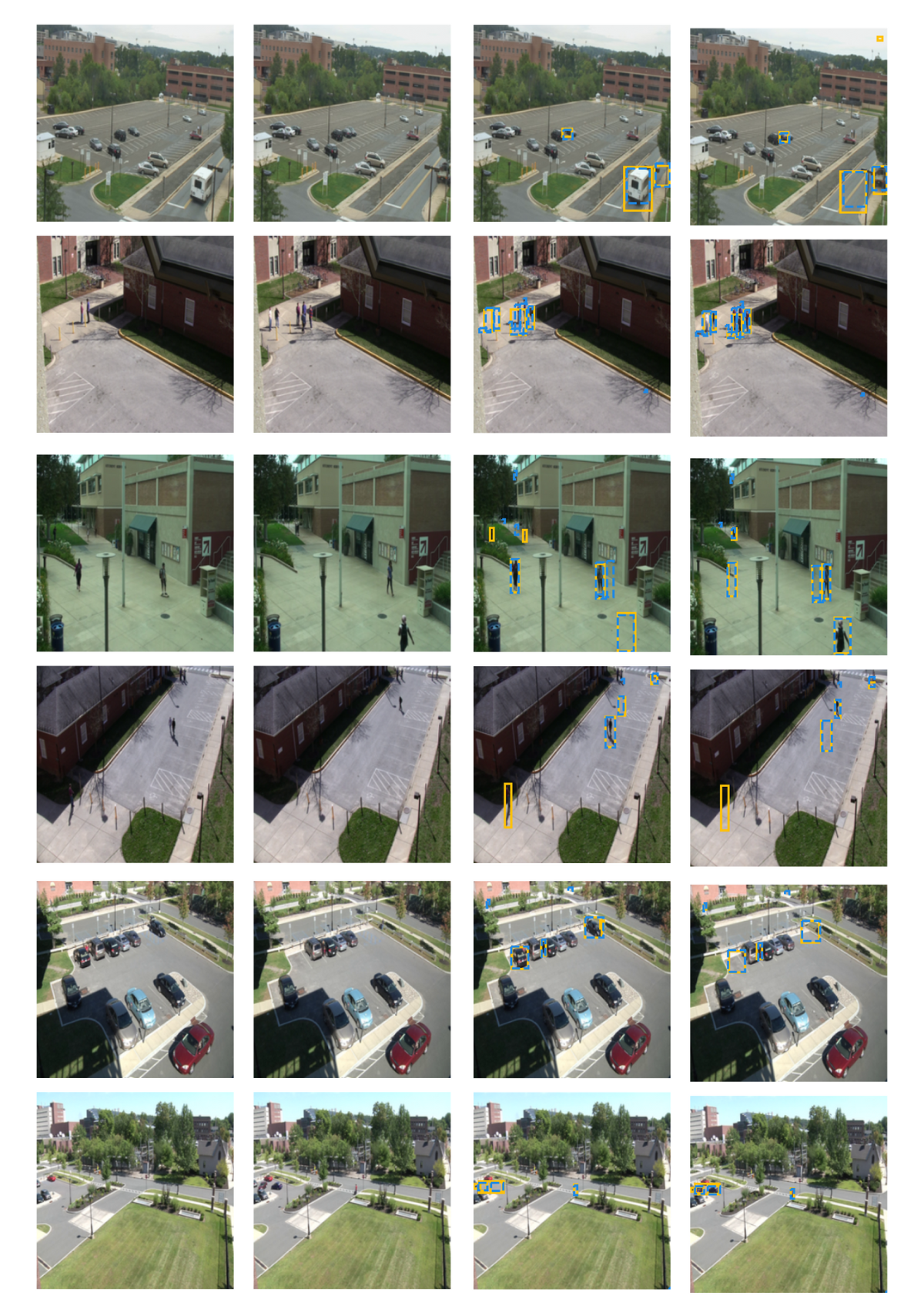}
    \caption{VIRAT-STD}
    \label{fig:coco}
\end{figure*}

\begin{figure*}[ht]
    \centering
    \includegraphics[width=0.9\linewidth]{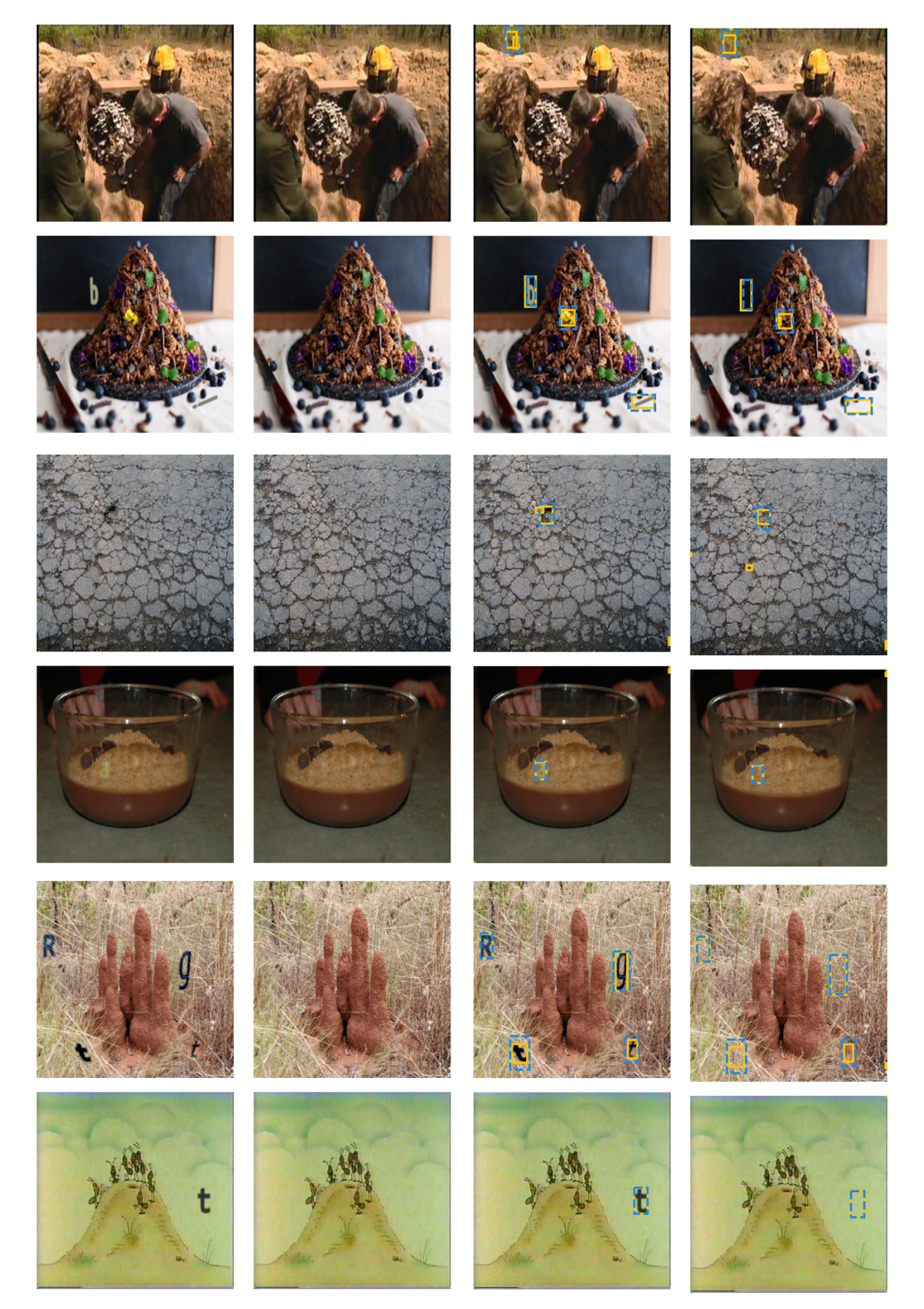}
    \caption{Synthtext-Change}
    \label{fig:coco}
\end{figure*}